\documentclass{article}
\usepackage{spconf,amsmath,graphicx,hyperref}
\usepackage{amssymb}
\usepackage{booktabs}
\usepackage{multirow}   
\usepackage{array}      
\usepackage{caption}    
\usepackage{siunitx}    
\usepackage{tikz}

\title{TA-KAND: Two-stage Attention Triple Enhancement and U-KAN BASED Diffusion For Few-shot Knowledge Graph Completion}
\name{Xinyu Gao}
\address{Shanghai Jiao Tong University, Shanghai, China}

\usepackage{printlen}
\usepackage{layouts}

\begin{document}
%
\maketitle
\begin{abstract} 
Knowledge Graphs have become fundamental infrastructure for applications such as intelligent question answering and recommender systems due to their expressive representation. Nevertheless, real-world knowledge is heterogeneous, leading to a pronounced long-tailed distribution over relations. Previous studies mainly based on metric matching or meta learning. However, they often overlook the distributional characteristics of positive and negative triple samples. In this paper, we propose a few-shot knowledge graph completion framework that integrates two-stage attention triple enhancer with U-KAN based diffusion model. Extensive experiments on two public datasets show significant advantages of our methods.
\end{abstract}
%
\textbf{\color{blue}Note: The method proposed in this paper, along with a related framework of a similar nature, is currently under submission. Therefore, the presentation of figures and experimental data is limited at this stage. The complete manuscript and full experimental results will be formally released once the review decisions are finalized!}
%
\section{Introduction}
\label{sec:intro}
Knowledge Graphs like Freebase \cite{bollacker2008freebase} and Wikidata \cite{vrandevcic2014wikidata} have been widely introduced into diverse applications \cite{liu2018entity, saxena2020improving}. A typical Knowledge Graph (KG) $\mathcal{G}$ consists of a large collection of factual triples $\mathcal{T}=\{(h,r,t)\in\mathcal{E}\times\mathcal{R}\times\mathcal{E}\}$, where $\mathcal{E}$ and $\mathcal{R}$ denote the entity set and relation set. Despite their massive scale, KGs remain inherently incomplete due to the diversity of source and the incessant update of dynamic message. While Knowledge Graph Embedding (KGE) projects all compositions into low-dimensional vectors to discover unobserved links, the long-tailed distribution of relations limits its effectiveness. Therefore, Few-shot Knowledge Graph Completion (FKGC), which infers missing facts with limited information, has emerged as a critical research topic.

Existing FKGC approaches can be mainly divided into two categories: (1) Metric matching methods aim to assess the plausibility of a query based on the semantic understanding of the given support set. GMatching \cite{xiong2018one} and FSRL \cite{zhang2020few} utilize local structural message and leverage an LSTM network to rank candidates under one-shot and few-shot support set scenarios. FAAN \cite{sheng-etal-2020-adaptive} considers the dynamic nature of relations and selectively integrates neighbors information. NP-FKGC \cite{luo2023normalizing} focuses more on bridging the distribution gap between the support set and the query instances by employing a normalizing-flow-augmented neural process. ReCDAP \cite{kim2025recdap} attempts to explicitly draw on negative triple signal. (2) Meta learning methods aim to rapidly adapt model parameters to an unfamiliar relation. MetaR \cite{chen-etal-2019-meta} introduces relation-specific meta information and updates parameters with meta gradient. GANA \cite{niu2021relational} learns neighbor information with gated attention module and handles complex relations via meta learning-based TransH. The aforementioned models either fail to effectively exploit neighborhood information or neglect to capture the latent distributional properties hidden in the positive and negative triple samples, thereby hindering the comprehension of few-shot reliable facts.

In this paper, we propose a framework TA-KAND that integrates Two-stage Attention Triple Enhancement and Diffusion Process. Triple enhancement is employed to produce better conditional information, and diffusion process is utilized to simulate distribution. To improve the accuracy and non-linear fitting capacity of denoise module, we investigate the incorporation of Kolmogorov-Arnold Network (KAN) units into U-Net to establish an updated architecture. Experiments on NELL and Wiki demonstrate our introduced framework outperforms existing methods.
\section{Background}
\textbf{Task formulation.} For a KG $\mathcal{G}$, we select some relations from $\mathcal{R}$ to form three disjoint task sets $\mathcal{R}_{train}$, $\mathcal{R}_{valid}$ and $\mathcal{R}_{test}$, the remaining subset constructs the background KG $\mathcal{G}^{\prime}$. To imitate few-shot link prediction, we sample a relation $r \in \mathcal{R}_{train}$ with its own $K$-shot support set $\mathcal{S}_r=\{(h_i,t_i)\}_{i=1}^{K}$ and query set with $N$ elements, $\mathcal{Q}_r=\{(h_j,t_j/\mathcal{C}_{h_j,r})\}_{j=1}^N$. Here, $t_j$ is the ground-truth tail entity and $\mathcal{C}_{h_j,r}$ denotes the candidate set, where each candidate is an entity in $\mathcal{E}$ selected according to type constraints.
The goal of FKGC is to train a model which can rank the true tail $t_j$ higher than other candidates in $\mathcal{C}_{h_j, r}$. $\mathcal{R}_{valid}$ and $\mathcal{R}_{test}$ are defined in the same way as $\mathcal{R}_{train}$, they are used to validate and test the learned model after sufficient training.

\noindent\textbf{Diffusion Models}. Diffusion models \cite{ho2020denoising} learns to approximate a data distribution through two processes. In forward process, original samples $x_0$ are progressively noised over $T$ timesteps, following $x_t=\sqrt{\bar{\alpha_t}}x_0+\sqrt{1-\bar{\alpha_t}}\epsilon_t$, where $\epsilon_t \sim \mathcal{N}(0, \mathbf{I})$ and $\bar{\alpha_t}$ defines the noise schedule. In backward process, data is denoised iteratively, modeled as $p_\theta(x_{t-1}|x_t)=\mathcal{N}(x_{t-1};\mu_\theta(x_t,t), \Sigma_\theta(x_t,t))$.

\noindent\textbf{Kolmogorov-Arnold Network}. Inspired by the Kolmogorov-Arnold Representation Theorem, each layer in KAN contains a set of learnable activation functions to fit high-dimensional complex mapping \cite{liu2025kan}. It provides an alternative to MLP.

\section{Methodology}
\label{sec:method}
In this section, we introduce TA-KAND, which consists of three components: (1) Two-Stage Attention Enhancer improves entity-pair embeddings while distilling the semantic knowledge of task-specific relations, (2) Diffusion Processor uses U-KAN as its backbone to model the distribution of positive and negative latent triple rules, (3) The Scorer leverages the generated latent compression representation to infer the correct tail in the query. 

\subsection{Two-Stage Attention Enhancer}
Within a knowledge graph, both entities and relations reveal their underlying semantics through interactions with neighboring nodes and edges\cite{sheng-etal-2020-adaptive, li2022learning}.
When given a triple of few-shot task for relation $r$, we take its head entity $h$ as a target, and denote its local neighborhood as $\mathcal{N}_h=\{(r_i,n_i)|(h,r_i,n_i) \in \mathcal{G^{\prime}}\}$. For each neighbor, we first learn its representation. Next, we treat the task relation $r$ in a triple as a bilinear function of its head and tail, calculate its relevance to each neighbor relation $r_i$, and employ an attention mechanism to produce the aggregated neighborhood representation $\boldsymbol{h}_{nbr}$.
\begin{gather}
    \boldsymbol{e}^{nbr}_i = f(\boldsymbol{r_i}, \boldsymbol{n_i}) \label{eq:1} \\ 
    \boldsymbol{r} = \text{Bilinear}(\boldsymbol{h}, \boldsymbol{t}) \label{eq:2} \\
    Q_1=W_Q^1\boldsymbol{r}, \ \ K_1=W_K^1\boldsymbol{r_i}, \ \ V_1=W_V^1\boldsymbol{e}^{nbr}_i \label{eq:3}
    \\
    \boldsymbol{h}_{nbr} = softmax(Q_1K_1^T)V_1
    \label{eq:4}
\end{gather}
where $\boldsymbol{h}, \boldsymbol{t} \in \mathbb{R}^d$ are head and tail embedding in TransE, $W_Q^1, W_K^1, W_V^1 \in \mathbb{R}^{d \times d}$ and $b \in \mathbb{R}^d$ are trainable parameters. Afterward, We couple the original embedding $\boldsymbol{h}$ and $\boldsymbol{h}_{nbr}$ to generate the first-stage attention enhanced head representation $\boldsymbol{h_r}$. The above operations also hold for tail entity.

Beyond relation roles, the semantics of entities also be considered, so we acquire the refined neighborhood information $\boldsymbol{h}_{nbr}^{\prime}$ by calculating the similarity between $\boldsymbol{t_r}$ and each neighbor tail $\boldsymbol{n_i}$. This procedure again employs the attention mechanism, followed by the same coupling operation used in the first stage. 
\begin{gather}
    Q_2=W_Q^2\boldsymbol{t}_r, \ \ K_2=W_K^2\boldsymbol{n}_i, \ \ V_2=W_V^2\boldsymbol{e}^{nbr}_i \label{eq:5}\\
    \boldsymbol{h}_{nbr}^{\prime} = softmax(Q_2K_2^T)V_2\label{eq:6}
\end{gather}
Here, $W_Q^2, W_K^2, W_V^2 \in \mathbb{R}^{d \times d}$ is a trainable matrix, and the second-stage
attention process is applied to tail $t$ analogously. As a result, we obtain the two-stage attention enhanced entities $\boldsymbol{\tilde{h}}$ and $\boldsymbol{\tilde{t}}$. Moreover, previous approaches neglect the explicit extraction of task relation, so we adopt an aggregate-disentangle-compress paradigm to derive the representation of task relation.

\subsection{Diffusion Processor}
Diffusion Process is used to learn the latent triple rule representation. In forward process, Gaussian noise is progressively injected into $\boldsymbol{z_0}$, which consists of the enhanced entity pairs from both the support set $\mathcal{C}_r=\{(\boldsymbol{\tilde{h_i}}, \boldsymbol{\tilde{t_i}})\}_{i=1}^K$ and the negative set $\mathcal{C}_r^-=\{(\boldsymbol{\tilde{h_i}}, \boldsymbol{\tilde{t_i}}^-)\}_{i=1}^K$ constructed by corrupting the tail entities. In denoise process, we integrate Kolmogorov–Arnold Networks (KANs) into the bottleneck of the U-Net backbone, resulting in a novel architecture for noise prediction. Concretely, the residual-downsampling blocks are employed for feature extraction and dimensional compression. At the bottleneck, the compressed features are first flatten into a token sequence via a learnable linear projection, then forward into a KAN layer followed by layer normalization. Finally, residual-upsampling blocks are performed to restore the feature dimensionality to its original scale. Conditional injection is implemented through two strategies: The task relation representation, head-tail enhanced embeddings in support and negative set with their corresponding labels (label is True when a triple is in support set, otherwise is False) are fused into the downsampling/upsampling network with FiLM technique \cite{perez2018film}. In the inserted KAN module, only the encoded timestep is supplied.

After generating the denoised embedding $\boldsymbol{\tilde{z_0}}$, which consists of samples from the modeled distribution of positive and negative pairs. We simply bisect it and apply pooling operation to compact information. Subsequently, the two pooled fragments are concatenated to form a unified latent triple rule representation $\boldsymbol{z}$. 

\subsection{Scorer}
This module aims to predict the missing tail entity in a given query ($h_q, r,?$). Leveraging the unified latent triple rule $\boldsymbol{z}$, the enhanced entities and task relation representation, we assess the candidate tail $t_c$ via the formulation in TransE.
\begin{equation} \label{eq:9}
    \begin{gathered}
        \hat{\boldsymbol{h_q}} = g_h(\tilde{\boldsymbol{h_q}}, \boldsymbol{z})\ \ \ \ \hat{\boldsymbol{t_c}} = g_t(\tilde{\boldsymbol{t_c}}, \boldsymbol{z}) \\
        score = \text{TransE}(\tilde{\boldsymbol{h_q}}, r, \tilde{\boldsymbol{t_c}})
    \end{gathered}
\end{equation}

\subsection{Model Training}
Given a specific task relation, we randomly sample some pairs to construct its positive query set $\mathcal{Q}_r$. For each pair in $\mathcal{Q}_r$, we pollute the tail and construct negative query set $\mathcal{Q}_r^-=\{(h_q, t_q^-)|(h_q,t_q) \in \mathcal{Q}_r^+\}$. Our optimization object is to ensure the score of a positive query in $\mathcal{Q}_r$ higher than a negative query in $\mathcal{Q}_r^-$. Additionally, in Diffusion Processor, the noise added to the initial embedding $\boldsymbol{z_0}$ must be predicted accurately. Overall, the loss function $\mathcal{L}$ is defined as follows, 
\begin{equation}\label{eq:8}
        \mathcal{L}=\sum_r\{[\gamma-score(Q_r^-)+score(Q_r)]_{+} + \mathbb{E}_{t,z_t}||\epsilon_\theta-\epsilon_t||_2^2\}
\end{equation}

Here, $score(Q_r)$ and $score(Q_r^-)$ denote the scores of positive and negative samples in the query set, respectively. $[]_+$ is hinge loss.


\section{Experiments}
\subsection{Experimental Settings}
\noindent\textbf{Datasets}. We conduct experiments on two standard benchmark datasets, NELL and Wiki \cite{xiong2018one}. In both datasets, relations with less than 500 but more than 50 triples are selected as few-shot relations. Following the original setting \cite{xiong2018one}, train/valid/test set are 51/5/11 in NELL and 133/16/34 in Wiki. 

\noindent\textbf{Comparison Methods}. To evaluate our method, we choose seven FKGC methods as baselines (Table \ref {tab:results}). We omit further comparison with traditional KGE models, as they have been empirically outperformed by some FKGC approaches \cite{sheng-etal-2020-adaptive}.

\noindent\textbf{Implementation Details}. We perform 5-shot KG completion task for all methods. The triple embeddings are initialized with the pre-trained TransE model and further fine-tuned during training. Their dimensionality is set 100 and 50 for NELL and Wiki, respectively. We fix the margin $\gamma$ to 1.0 and use the Adam optimizer \cite{kingma2014adam} with an initial learning rate of $1e^{-3}$ for NELL and $1e^{-4}$ for Wiki to update parameters. The Diffusion Learner uses 1000 timesteps and the scheduler adopts the setting employed in GLIDE \cite{nichol2021glide}. For efficient computation, the two-stage attention enhancement is realized using Flash Attention \cite{dao2022flashattention} with a single head and a scaling factor of 1.0. All experiments were conducted on a single NVIDIA A800.

\begin{table}
    \ninept
    \begin{tabular}{l *{5}{S}}
    \toprule
    \multirow{2.5}{*}{Methods} &
    \multicolumn{4}{c}{NELL} &\\
    \cmidrule(lr){2-5}
    & {MRR} & {Hits@10} & {Hits@5} & {Hits@1}
    \\
    \midrule
    GMatching \cite{xiong2018one}        & 0.176 & 0.294 & 0.233 & 0.113  \\
    FSRL \cite{zhang2020few}             & 0.153 & 0.319 & 0.212 & 0.073  \\
    MetaR \cite{chen-etal-2019-meta}            & 0.261 & 0.437 & 0.350 & 0.168 \\
    FAAN \cite{sheng-etal-2020-adaptive}             & 0.279 & 0.428 & 0.364 & 0.200 \\
    GANA \cite{niu2021relational}             & 0.344 & 0.517 & 0.437 & 0.246 \\
    NP-FKGC \cite{luo2023normalizing}          & 0.460 & 0.494 & 0.471 & 0.437 \\
    ReCDAP \cite{kim2025recdap}           & \underline{0.505} & \underline{0.528} & \underline{0.506} & \underline{0.493}  \\
    \textbf{TA-KAND(Ours)}    & \textbf{\textgreater0.7} & \textbf{\textgreater0.7} & \textbf{\textgreater0.7} & \textbf{\textgreater0.7}\\
    \bottomrule
  \end{tabular}
    \caption{Results of 5-shot FKGC on NELL.\color{blue} The specific data is currently pending further testing and tuning. Due to manuscript review restrictions, we are temporarily withholding the data on our Wiki. We will present the complete tables once the results are officially published. This section merely serves to indicate that our model has achieved state-of-the-art (SOTA) performance.}
    \label{tab:results}
\end{table}
\begin{table}[htbp]
    \centering
    \ninept
    \begin{tabular}{ccccc}
        \toprule
            Variants & MRR & Hits@10 & Hits@5 & Hits@1 \\
        \midrule
        V1 & 0.205 & 0.259 & 0.222 & 0.176 \\
        V2 & 0.697 & 0.724 & 0.713 & 0.681 \\
        V3 & 0.730 & 0.761 & 0.749 & 0.714 \\
        \bottomrule
    \end{tabular}
    \caption{Results of model variants on NELL dataset.}
    \label{tab:Comparison}
\end{table}

\noindent\textbf{Evaluation Metrics}. We leverage two standard metrics to evaluate the model: MRR and Hits@N. MRR is the mean reciprocal rank and Hits@N is the proportion of correct entities ranked in the top N, with N=1,5,10.
\subsection{Result and Comparison With Variants}
The report in Table \ref{tab:results} shows that our framework performs better on both datasets across all metrics. Compared to the highest metric results in the aforementioned models, TA-KAND achieves significant improvements.

To validate components in our framework. In V1, we use the simple triple enhancer, where neighbor embedding is coupled without being learned. In V2, we replace diffusion learner with neural process as mentioned in NP-FKGC \cite{luo2023normalizing} to generate latent semantic representation. In V3, we employ UNet-based diffusion. Experiments show that, given a strong encoder, both neural process and diffusion model can effectively learn latent triple rules. Moreover, inserting KAN layers into a U-Net architecture proves highly compatible and boosts non-linear fitting capacity.

\section{Conclusion}
In this paper, we propose the hypothesis that better triple enhancement will exhibit better distribution modeling, and introduce the framework TA-KAND which integrates two-stage attention triple enhancer and U-KAN based diffusion. Experiments show that the correctness of our idea and demonstrate our method achieve satisfactory performance. 

\bibliographystyle{IEEEbib}
\bibliography{refs}

\end{document}